# Inference Algorithms for Similarity Networks

**Dan Geiger**
Department of Computer Science
Technion Israel Institute of Technology
Technion City, Haifa Israel 32000
<dang@cs.technion.ac.il>

**David Heckerman**
Microsoft Research Center and
Department of Computer Science, UCLA
One Microsoft Way, 9S/1024
Redmond, WA 98052-6399
<heckerma@microsoft.com>

**Abstract**

We examine two types of similarity networks each based on a distinct notion of relevance. For both types of similarity networks we present an efficient inference algorithm that works under the assumption that every event has a nonzero probability of occurrence. Another inference algorithm is developed for type 1 similarity networks that works under no restriction, albeit less efficiently.

## 1 INTRODUCTION

Similarity networks were invented by Heckerman (1991) as a tool for constructing large Bayesian networks from the judgments of domain experts. Heckerman used them to construct a diagnostic system for lymph-node pathology. The main advantages of similarity networks are their ability to improve the quality of the domain expert's judgments and to utilize statements of conditional independence that are not represented in a Bayesian network in order to reduce the number of probabilities the expert needs to specify.

In (Geiger and Heckerman, 1991), we removed several technical restrictions imposed by the original development, and showed how to use a similarity network directly for inference without converting it to a Bayesian network as proposed in (Heckerman, 1991). In do so, we showed how to take advantage of asymmetric independence assertions to speed up inference.

In this paper, we define two types of similarity networks each based on a distinct notion of relevance. We specify more fully the inference algorithm outlined in (Geiger and Heckerman, 1991), and prove that it works for type 1 similarity networks. We also develop a faster inference algorithm that works for both types of similarity networks, under the assumption that every event has a nonzero probability of occurrence.

We assume the reader is familiar with the definition and usage of Bayesian networks. For details consult (Pearl, 1988).

## 2 DEFINITION OF SIMILARITY NETWORKS

Consider a mutually exclusive and exhaustive set of hypotheses such as the list of possible identifications in a classification task. We will represent these hypotheses as *values* of a variable called $h$. The variable $h$ is the focus of construction for similarity networks, and sometimes is called the *distinguished* variable or *hypothesis* variable. We refer to other variables in a given domain as *nondistinguished* variables. Each value of $h$ is called a *hypothesis*.

**Definition** A *cover* of a set of hypotheses $H$ is a collection $\{A_1, \ldots, A_k\}$ of nonempty subsets of $H$ whose union is $H$. Each cover is a hypergraph, called a *similarity hypergraph*, where the $A_i$ are hyperedges and the hypotheses are nodes. A cover is *connected* if the similarity hypergraph is connected.

A similarity network is a set of Bayesian networks, each constructed under the assumption that $h$ draws its value from a specific subset of its domain. The event "$h$ draws its value from the set $A_i$" is denoted by $[\![A_i]\!]$.

**Definition** Let $P(h, u_1, \ldots, u_n)$ be a probability distribution and $A_1, \ldots, A_k$ be a connected cover of the values of $h$. A directed acyclic graph $D_i$ is called a *local network of $P$ associated with $A_i$* if $D_i$ is a Bayesian network of $P(h, v_1, \ldots, v_m \mid [\![A_i]\!])$, where $\{v_1, \ldots, v_m\}$ is the set of all variables in $\{u_1, \ldots, u_n\}$ that "help to discriminate" the hypotheses in $A_i$. The set of $k$ local networks is called a *similarity network* of $P$.

We define "help to discriminate" formally in the next section.

Let us examine this definition, using the following example taken from (Geiger and Heckerman, 1991):

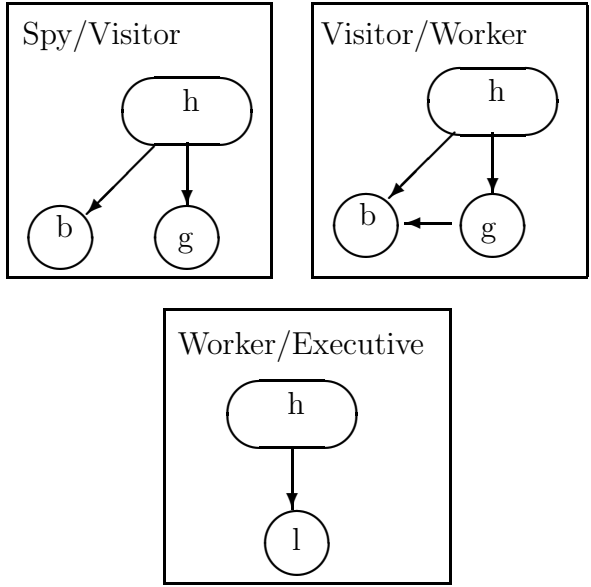

Figure 1: A similarity network representation of the secured-building story.

> A guard of a secured building expects four types of persons to approach the building's entrance: executives, regular workers, approved visitors, and spies. As a person approaches the building, the guard can note its gender ($g$), whether or not the person wears a badge ($b$), and whether or not the person arrives in a limousine ($l$). We assume that only executives arrive in limousines and that male and female executives wear badges just as do regular workers (to serve as role models). Furthermore, we assume that spies are mostly men. Spies always wear badges in an attempt to fool the guard. Visitors don't wear badges because they don't have one. Female-workers tend to wear badges more often than do male-workers. The task of the guard is to identify the type of person approaching the building.

This problem is represented by the similarity network shown in Figure 1. The similarity network in this figure is based on the cover {*spy, visitor*}, {*visitor, worker*}, {*worker, executive*} of the hypotheses set. This cover is connected, because it consists of the three links *spy—visitor—worker—executive* which form a connected hypergraph. This similarity network contains three local networks: one local network helps to discriminate spies from visitors, another local network helps to discriminate visitors from workers, and a third local network helps to discriminate workers from executives.

In each local network, we include only those variables that help to discriminate among the hypotheses covered by that local network. For example, badge worn and gender are not included in the local network for workers versus executives, because they do not help to distinguish between these hypotheses. Similarly, the variable representing arrival in a limousine is not included in the local networks for spies versus visitors or visitors versus workers.

The relationship between gender and the hypothesis variable $h$ is an example of *subset independence*, whereby a nondistinguished variable is independent of $h$, given $h$ draws its values from a subset of hypotheses {*worker, executive*}. In general, an assertion of subset independence is represented in a similarity network whenever a link between the hypothesis node and a nondistinguished variable exists in some local networks, but not in all networks. The relationship between badge worn and $h$, when $h$ is restricted to {*worker, executive*}, and the relationship of arrival in limousine and $h$, when $h$ is restricted to {*spy, visitor*}, are additional examples of subset independence.

The relationship between gender and badge worn is an example of *hypothesis-specific* independence, whereby two variables are independent given some hypotheses { *spies, visitors*}, but dependent given others {*visitors, workers*}. In general, a hypothesis-specific independence assertion is represented in a similarity network whenever a link between two nondistinguished variables exists in some local networks, but does not exists in other local networks.[1]

The definition of similarity networks does not specify how to select a connected cover of hypotheses. Although any selection of a connected cover yields a valid similarity network, some selections yield similarity networks that display more subset independence assertions than do other selections. An analogous situation exists when constructing a Bayesian network where some construction orders yield Bayesian networks that display more symmetric independence assertions than do other Bayesian networks. The practical solution for constructing a Bayesian network is to choose a construction order according to cause-effect relationships. This selection tends to maximize the information about symmetric independence encoded in the resulting network. The practical solution for constructing the similarity hypergraph is to choose a connected cover by grouping together hypotheses that are "similar" to each other by some criteria under our control (e.g., spies and visitors). This choice tends to maximize the number of subset independence assertions encoded in a similarity network. Hence the name for this representation.

---

[1] Heckerman (1991) coined the terms subset independence and hypothesis-specific independence.

# 3 TWO TYPES OF SIMILARITY NETWORKS

The definition of similarity networks is not complete without attributing a precise meaning to the utterance "helps to discriminate" used in the definition of a local network. We give two possibilities based on the study of three relations *coupled*, *related*, and *relevant* presented in (Geiger and Heckerman, 1990).

**Definition** Let $P(u_1, \ldots, u_n \mid \mathbf{e})$ be a probability distribution where $\mathbf{e}$ is a fixed event. Variables $u_i$ and $u_j$ are *unrelated given* $\mathbf{e}$, if $u_i$ and $u_j$ are disconnected in every minimal Bayesian network of $P(u_1, \ldots, u_n \mid \mathbf{e})$. Otherwise, $u_i$ and $u_j$ are *related given* $\mathbf{e}$, denoted $related(u_i, u_j \mid \mathbf{e})$.

This definition states that two variables $u_i$ an $u_j$ are unrelated given $\mathbf{e}$, if there exists no trail connecting them—that is, there exists no sequence of variables $u_i, \ldots, u_j$ such that every two consecutive variables in this sequence are connected with a link. The requirement that $u_i$ and $u_j$ be disconnected in *every* minimal network is not as strong as it may seem, because if $u_i$ and $u_j$ are disconnected in one minimal Bayesian network of $P$, then $u_i$ and $u_j$ are disconnected in every minimal Bayesian network of $P$ (Geiger and Heckerman, 1990).

**Definition** Let $P(u_1, \ldots, u_n \mid \mathbf{e})$ be a probability distribution where $\mathbf{e}$ is a fixed event. Variables $u_i$ and $u_j$ are *mutually irrelevant given* $\mathbf{e}$, if $P(u_i \mid u_j, v_1, \ldots, v_m, \mathbf{e}) = P(u_i \mid v_1, \ldots, v_m, \mathbf{e})$ or $P(v_1, \ldots, v_m \mid \mathbf{e}) = 0$ for every value of $v_1, \ldots, v_m$, where $\{v_1, \ldots, v_m\}$ is an arbitrary subset of $\{u_1, \ldots, u_n\} \setminus \{u_i, u_j\}$. Otherwise, $u_i$ and $u_j$ are *mutually relevant given* $\mathbf{e}$, denoted $relevant(u_i, u_j \mid \mathbf{e})$.

This definition states that two variables $u_i$ and $u_j$ are mutually irrelevant given $\mathbf{e}$ if for any consistent assignment for the variables of interest knowing the value of one variable does not change the knowledge about the values of the other.

To use the concepts of relatedness and relevance for defining those variables that are included in a local network, we distinguish one variable as the hypothesis variable (call it $h$), and define the event $\mathbf{e}$ to be $[\![A_i]\!]$—namely, a disjunction over a subset of the values of $h$.

If $\mathbf{e}$ is the event $[\![A]\!]$, where $A$ is the set of *all* values of $h$—that is, $\mathbf{e}$ states that variable $h$ is assigned one of its values—then instead of $related(u_i, u_j \mid [\![A]\!])$, we write $related(u_i, u_j)$ and say that $u_i$ and $u_j$ are *related*. Similarly, instead of $relevant(u_i, u_j \mid [\![A]\!])$ we write $relevant(u_i, u_j)$ and say that $u_i$ and $u_j$ are mutually relevant.

**Definition** A similarity network constructed by including in each local network $D_i$ only those variables $u$ that satisfy $related(u, h \mid [\![A_i]\!])$ is said to be of *type 1*.

A similarity network constructed by including in each local network $D_i$ only those variables $u$ that satisfy $relevant(u, h \mid [\![A_i]\!])$ is said to be of *type 2*.

An equivalent definition for type 1 similarity networks is entailed by the following theorem:

**Theorem 1 (Geiger and Heckerman, 1990)** *Let $P(u_1, \ldots, u_n \mid \mathbf{e})$ be a probability distribution where $U = \{u_1, \ldots, u_n\}$ and $\mathbf{e}$ be a fixed event. Then, $u_i$ and $u_j$ are* unrelated given $\mathbf{e}$ *iff there exist a partition $U_1, U_2$ of $U$ such that $u_i \in U_1$, $u_j \in U_2$, and $P(U_1, U_2 \mid \mathbf{e}) = P(U_1 \mid \mathbf{e})P(U_2 \mid \mathbf{e})$.* [2]

We can now associate relatedness and relevance.

**Theorem 2** *Let $P(u_1, \ldots, u_n \mid \mathbf{e})$ be a probability distribution where $\mathbf{e}$ is a fixed event. Then, for every $u_i$ and $u_j$, $relevant(u_i, u_j \mid \mathbf{e})$ implies $related(u_i, u_j \mid \mathbf{e})$.*

**Proof:** Suppose $u_i$ and $u_j$ are not related given $\mathbf{e}$. Let $U_1, U_2$ be a partition of $U$ such that $u_i \in U_1$, $u_j \in U_2$ and $I(U_1, U_2 \mid \emptyset)$ holds for $P$. Theorem 1 guarantees the existence of these conditions. We show that $u_i$ and $u_j$ must be mutually irrelevant given $\mathbf{e}$. Let $Z$ be an arbitrary subset of $\{u_1, \ldots, u_n\} \setminus \{u_i, u_j\}$. Let $Z_1 = Z \cap U_1$ and $Z_2 = Z \cap U_2$. Since $I(U_1, U_2 \mid \emptyset)$ holds in $P(u_1, \ldots, u_n \mid \mathbf{e})$, it follows, by summing over the variables not in $\{u_i, u_j\} \cup Z_1 \cup Z_2$, that $I(\{u_i\} \cup Z_1, \{u_j\} \cup Z_2 \mid \emptyset)$ holds. Consequently, $I(\{u_i\}, \{u_j\} \mid Z_1 \cup Z_2)$ must hold as well (Pearl 1988, Equation 3.6c). Thus, $I(\{u_i\}, \{u_j\} \mid Z)$ holds for every $Z$. Hence, $u_i$ and $u_j$ are mutually irrelevant given $\mathbf{e}$. (This proof is a slight modification of the one appearing in (Geiger and Heckerman, 1990)). □

In particular, Theorem 2 holds when $\mathbf{e}$ is the event $[\![A_i]\!]$ where $A_i$ is a subset of values of some variable in $U$. Consequently, a type 2 similarity network of $P$ always includes in each local network at least all the variables included in that local network by a type 1 similarity network. Moreover, whenever a variable $u$ does not satisfy $related(u, h)$, it will not be included in any local network of a type 1 or a type 2 similarity network.

It is not hard to construct an example showing that *related* does not imply *relevant*. For example, consider a Markov chain $P(x, y, z) = P(x)P(y \mid x)P(z \mid y)$ where $x$ and $z$ are binary variables, and $y$ is a trinary variable. Any two matrices $P(y \mid x)$ and $P(z \mid y)$ whose lines and columns are linearly independent and whose product yields a matrix with identical lines will render $x$ and $z$ (marginally) independent. Since $x$ and $z$ are also independent given $y$, it follows that $x$ and $z$ are mutually irrelevant, although $x$ and $z$ are related.

A necessary and sufficient condition for the converse of Theorem 2 to hold is that $P(u_1, \ldots, u_n \mid \mathbf{e})$ be *tran-*

---

[2] In (Geiger and Heckerman, 1990), $P(U \mid \mathbf{e})$ is replaced with $P(U)$. Since $\mathbf{e}$ is a fixed event, this shift of notation does not alter the proof of this theorem.

*sitive.* Namely, for every three variables $u_i, u_j$ and $u_k$,

$$\begin{aligned} &relevant(u_i, u_j \mid \mathbf{e}) \ \& \ relevant(u_j, u_k \mid \mathbf{e}) \\ &\Rightarrow relevant(u_i, u_k \mid \mathbf{e}) \end{aligned} \quad (1)$$

(Geiger and Heckerman, 1990).

Indeed, in the previous example, transitivity is violated, because $x$ and $y$ are mutually relevant and $y$ and $z$ are mutually relevant, but $x$ and $z$ are not mutually relevant. Notably, the relation *related* is always transitive, because connectivity in graphs is transitive.

## 4  INFERENCE USING SIMILARITY NETWORKS

The main task similarity networks are designed for is to compute the posterior probability of each hypothesis given a set of observations, as is the case in diagnosis. In this section, we show that under reasonable assumptions, the computation of the posterior probability of each hypothesis can be done in each local network and then be combined coherently according to the axioms of probability theory. We analyze the complexity of our algorithm demonstrating its superiority over inference algorithms that operate on Bayesian networks.

We assume that any instantiation of the variables in a similarity network of $P$ has a nonzero probability of occurrence. Such a probability distribution is said to be *strictly positive*. This assumption can be reasonable in such applications as medical diagnosis, where given an arbitrary collection of clinical findings, each potential disease retains some probability of occurrence. Subject to this assumption, we develop an inference algorithm that operates both on type 1 and type 2 similarity networks. We will remove this assumption latter at the cost of obtaining an inference algorithm that operates only on type 1 similarity networks and whose complexity is higher.

The inference problem at hand can be stated as follows: Given a similarity network of $P(h, u_1, \ldots, u_n)$ that is based on a partition $\mathcal{A} = \{A_1, \ldots, A_k\}$ of the values of $h$, and given a set of assignments $\mathbf{v_1}, \ldots, \mathbf{v_m}$ for a set $v_1, \ldots, v_m$ of variables that is a subset of $\{u_1, \ldots, u_n\}$ compute $P(\mathbf{h_j} \mid \mathbf{v_1}, \ldots, \mathbf{v_m})$— the posterior probability of $\mathbf{h}_j$—for every $\mathbf{h}_j$.

In order to compute the posterior probability of each $\mathbf{h}_j$ we use the procedure *INFER*. This procedure has two parameters, one specifying a query of the form "compute $P(X \mid \mathbf{Y})$" and the second is a Bayesian network where $X$ and $Y$ are sets of variables that appear in the network and $\mathbf{Y}$ is a value of $Y$. We do not need to specify *INFER*'s operational details in order to demonstrate how this procedure is extended to operate on similarity networks. We now describe this new inference algorithm.

First, for each $\mathbf{h}_i$ we identify a set of hypotheses $A_j \in \mathcal{A}$ to which $\mathbf{h}_i$ belongs and compute the posterior probability of hypothesis $\mathbf{h}_i$ under the additional assumption that one of the hypotheses in $A_j$ holds true. In other words, we compute $P(\mathbf{h}_i \mid \mathbf{v_1}, \ldots, \mathbf{v_m}, [\![A_j]\!])$. Second, the posterior probabilities $P(\mathbf{h}_j \mid \mathbf{v_1}, \ldots, \mathbf{v_m})$ are computed from the probabilities $P(\mathbf{h}_j \mid \mathbf{v_1}, \ldots, \mathbf{v_m}, [\![A_i]\!])$ by solving a set of linear equations:

$$\begin{aligned} P(\mathbf{h}_j \mid \mathbf{v_1}, \ldots, \mathbf{v_m}) &= P(\mathbf{h}_j \mid \mathbf{v_1}, \ldots, \mathbf{v_m}, [\![A_i]\!]) \cdot \\ &\sum_{\mathbf{h}_j \in A_i} P(\mathbf{h}_j \mid \mathbf{v_1}, \ldots, \mathbf{v_m}) \end{aligned}$$

that relate these quantities. We shall see later that these equations have a unique solution.

It remains to show how to compute the query $P(\mathbf{h}_i \mid \mathbf{v_1}, \ldots, \mathbf{v_m}, [\![A_j]\!])$. It seems that one can merely call the procedure *INFER* to compute this query using the local network $D_j$ which corresponds to $A_j$. The query $P(\mathbf{h}_i \mid \mathbf{v_1}, \ldots, \mathbf{v_m}, [\![A_j]\!])$, however, may include variables that do not appear in $D_j$ in which case *INFER* is not applicable.

Fortunately, for both type 1 and for type 2 similarity networks the following equality will be shown to hold:

$$P(\mathbf{h}_i \mid \mathbf{v_1}, \ldots, \mathbf{v_l}, [\![A_j]\!]) = P(\mathbf{h}_i \mid \mathbf{v_1}, \ldots, \mathbf{v_m}, [\![A_j]\!]) \quad (2)$$

where $v_1, \ldots, v_l$ are the variables in $\{v_1, \ldots, v_m\}$ that appear in $D_j$ and $\mathbf{v_1}, \ldots, \mathbf{v_l}$ are their values. Thus to compute $P(\mathbf{h}_i \mid \mathbf{v_1}, \ldots, \mathbf{v_m}, [\![A_j]\!])$ we use the procedure *INFER* to compute the query $P(\mathbf{h}_i \mid \mathbf{v_1}, \ldots, \mathbf{v_l}, [\![A_j]\!])$ using the network $D_j$. Equation 2 tells us that the two computations yield identical answers.

For type 2 similarity networks, the justification of Equation 2 is that $v_{l+1}, \ldots, v_m$ are conditionally independent of $\mathbf{h}_i$ given every value of the variables $v_1, \ldots, v_l$ that appear in $D_j$ where $v_{l+1}, \ldots, v_m$ are the variables in $\{v_1, \ldots, v_m\}$ that do not appear in $D_j$. If Equation 2 does not hold, some of the variables in $\{v_{l+1}, \ldots, v_m\}$ would have appeared in the local network $D_j$, contrary to our assumption that $D_j$ contains only $v_1, \ldots, v_l$. Moreover, if some variable appears in a type 2 similarity network, then it will also appear in a type 1 similarity network (see the comment after Theorem 2). Therefore, this equality holds for type 1 similarity networks as well.

This algorithm is summarized below.

**Algorithm (Inference in similarity networks)**
**Input:** A similarity network of $P(u_1, \ldots, u_n, h)$ based on a connected cover $A_1, \ldots, A_k$ of $h$'s values.

**Output:** $P(h \mid \mathbf{v_1}, \ldots, \mathbf{v_m})$ where $\mathbf{v_1}, \ldots, \mathbf{v_m}$ are values of variables $v_1, \ldots, v_m$ and $\{v_1, \ldots, v_m\}$ is a subset of $\{u_1, \ldots, u_n\}$.

**Notation:** $D_j$ denotes the local network that corresponds to $A_j$ and $V_j$ are the variables that appear in $D_j$.

| | |
|---|---|
| 1 | For each $A_j$ |
| 2 | Let $\{v_1, \ldots, v_l\}$ be the variables in |

```
                V_j ∩ {v_1, ..., v_m}
3       For each h_i ∈ A_j
4           α_ij := INFER(P(h_i | v_1, ... v_l, [[A_j]]), D_j)
5           If α_ij = 0, then
                Return "P is not strictly positive"
6       Solve the following set of linear equations:
7           For all i and j, P(h_i | v_1, ..., v_m) =
                α_ij · Σ_{h_i ∈ A_j} P(h_i | v_1, ..., v_m)
8           Σ_i P(h_i | v_1, ..., v_m) = 1
9       Return P(h | v_1, ..., v_m)
```

We have argued already that the solution to the equations listed in Lines 7 and 8 provides the desired posterior probability. It remains to show that there exists a unique solution. Let us examine a local network $D_j$ that corresponds to $A_j$. Assume $A_j$ consists of $\mathbf{h_1}, \ldots, \mathbf{h_r}$. Since $\mathbf{v_1}, \ldots, \mathbf{v_m}$ remain fixed throughout the computations we denote $P(\mathbf{h_i} | \mathbf{v_1}, \ldots, \mathbf{v_m})$ by $Q(\mathbf{h_i})$. Consider the following equations:

$$Q(\mathbf{h_1}) = \alpha_{1,j} [Q(\mathbf{h_1}) + Q(\mathbf{h_2}) + \ldots + Q(\mathbf{h_r})] \quad (3)$$
$$Q(\mathbf{h_2}) = \alpha_{2,j} [Q(\mathbf{h_1}) + Q(\mathbf{h_2}) + \ldots + Q(\mathbf{h_r})] \quad (4)$$
$$\ldots$$
$$Q(\mathbf{h_r}) = \alpha_{r,j} [Q(\mathbf{h_1}) + Q(\mathbf{h_2}) + \ldots + Q(\mathbf{h_r})] \quad (5)$$

These are the subset of the equations defined in Line 7 which correspond to the local network $D_j$. By dividing every pair of consecutive equations, we obtain the following ratios:

$$\begin{aligned}
Q(\mathbf{h_r}) &= \frac{\alpha_{r,j}}{\alpha_{r-1,j}} Q(\mathbf{h_{r-1}}) \quad (6) \\
Q(\mathbf{h_{r-1}}) &= \frac{\alpha_{r-1,j}}{\alpha_{r-2,j}} Q(\mathbf{h_{r-2}}) \\
&\ldots \\
Q(\mathbf{h_2}) &= \frac{\alpha_{2,j}}{\alpha_{1,j}} Q(\mathbf{h_1})
\end{aligned}$$

Hence, the solution of these equations provides the ratios of the posterior probabilities between every pair of hypotheses in $A_j$. Since we repeat this process for every $A_j$ and since the cover defined by $A_1, \ldots A_k$ is connected, the ratio of every pair of hypotheses is established. To obtain the absolute values of each $Q(\mathbf{h_i})$, it remains to normalize their sum to one, using the Equation on Line 7 of the algorithm.

Consequently we have proven the following theorem.

**Theorem 3** *Let $P(h, u_1, \ldots, u_n)$ be a probability distribution and $\mathcal{A} = \{A_1, \ldots, A_k\}$ be a partition of the values of $h$. Let $S$ be a similarity network based on $\mathcal{A}$. Let $v_1, \ldots, v_m$ be a subset of variables whose value is given. There exists a single solution for the set of equations defined by Line 7 and 8 of the above algorithm and this solution determines uniquely the conditional probability $P(h | v_1, \ldots, v_m)$.*

An important observation to make is that the equations on Lines 7 and 8 are derived from a given probability distribution $P(h, u_1, \ldots, u_n)$. Consequently, although some equations might be redundant, these equations are always consistent. When the set of local networks is constructed from expert judgments, as done in practice, consistency is not guaranteed. Heckerman (1991) describes an algorithm that helps a user to construct a consistent set of local networks by prompting to his attention all probabilities that have already been assigned previously in another local network and verifying with him that these probabilities are acceptable.

It remains to analyze the complexity of this inference algorithm. For simplicity, we assume that all variables are binary in which case the procedure *INFER* has a complexity of $\mathcal{O}(2^n)$. In the worst case, the proposed inference algorithm may not perform more efficiently, because all $n$ variables may appear in each local network. In practice, however, each local network contains a small percentage, say c, of the $n$ variables, because all other variables are irrelevant given the context of a specific local network.[3] If $\mathcal{O}(n)$ local networks are given, the complexity of applying *INFER* to these local networks is $\mathcal{O}(n \cdot 2^{cn})$, which is smaller than $\mathcal{O}(2^n)$ obtained by applying *INFER* on a single Bayesian network generated from these local networks. The complexity of solving the equations on Line 7 and 8 is ignored, because it is linear in $n$. Thus from $2^{100}$ calculations, for example, we reduce to $100 \cdot 2^{20}$.

## 5 INFERENTIAL AND DIAGNOSTIC COMPLETENESS

An important property of Bayesian networks is that their parameters encode the entire joint distribution through the product rule. This property guarantees that any inference task can in principle be computed from the parameters encoded in a Bayesian network. Motivated by this observation we establish the following definition.

**Definition** A similarity network $S$ for $P(u_1, \ldots, u_n, h)$ is *inferentially complete* if the distribution $P(u_1, \ldots, u_n, h)$ can be recovered from the parameters of $S$.

Clearly not all similarity networks are inferentially complete. For example if $P(u_1, \ldots, u_n, h)$ factors into the product $P(u_1)P(u_2 \ldots, u_n, h)$ then the variable $u_1$ will not be included in any local network. Therefore, it will be impossible to recover $P(u_1)$ from the parameters encoded in the similarity networks of $P$. The information about $P(u_1)$ that is lost in the process of producing a similarity network of $P$, however, is never needed in order to compute the posterior probability of any hypothesis. Evidently, inferential completeness is too strong a requirement for the purpose of computing the posterior probability of each hypothesis.

---

[3] A reasonable number for c in the lymph-node-pathology domain is 0.2.

**Definition** A similarity network $S$ for $P(h, u_1, \ldots, u_n)$ is *diagnostically complete* if the conditional distribution $P(h \mid v_1, \ldots, v_m)$ can be recovered from the parameters of $S$ for every subset $\{v_1, \ldots, v_m\}$ of $\{u_1, \ldots, u_n\}$.

In the previous section, we showed that every type 1 or type 2 similarity network of a strictly positive probability distribution $P$ is diagnostically complete (Theorem 5). The inference algorithm we presented shows how to compute $P(h \mid \mathbf{v_1}, \ldots, \mathbf{v_m})$ for every value of $v_1, \ldots, v_m$. If $P$ is not strictly positive, one can construct examples where the equations defined by Lines 7 and 8 of our inference algorithm do not have a single solution. Nevertheless, we will prove that, under minor restrictions, every type 1 similarity network is diagnostically complete. We conjecture that every type 2 similarity network is diagnostically complete, under the same restrictions.

Before proving diagnostic completeness we resort to an example where our inference algorithm fails, and examine how the posterior probability can be computed in an alternative way. This computation highlights the general approach. Suppose $S$ is a similarity network for $P(h, y)$ where $h$ has three values $\{\mathbf{h}_1, \mathbf{h}_2, \mathbf{h}_3\}$ having equal apriori probability; and suppose that $y$ has two values $+y, -y$. Also assume that $S$ is based on the cover $\{\{\mathbf{h}_1, \mathbf{h}_2\}, \{\mathbf{h}_2, \mathbf{h}_3\}\}$, and that $P(+y \mid \mathbf{h}_2) = 0$.

When we apply our algorithm to compute $P(\mathbf{h}_i \mid +y)$, the algorithm generates three equations $P(\mathbf{h}_1 \mid +y, [\![\mathbf{h}_1, \mathbf{h}_2]\!]) = 1$, $P(\mathbf{h}_2 \mid +y, [\![\mathbf{h}_2, \mathbf{h}_3]\!]) = 0$, and $P(\mathbf{h}_3 \mid +y, [\![\mathbf{h}_2, \mathbf{h}_3]\!]) = 1$. From these three equations, we cannot compute the relative magnitude of the posterior probability of $\mathbf{h}_1$ versus $\mathbf{h}_3$. All three equations merely show that $P(\mathbf{h}_2 \mid +y)$ is zero.

$P(\mathbf{h}_i \mid +y)$, however, can be computed from the parameters that quantify $S$. These parameters include the following: $P(\mathbf{h}_1 \mid \mathbf{h}_1 \vee \mathbf{h}_2)$, $P(\mathbf{h}_2 \mid \mathbf{h}_2 \vee \mathbf{h}_3)$, $P(\mathbf{h}_3 \mid \mathbf{h}_2 \vee \mathbf{h}_3)$, and $P(+y \mid \mathbf{h}_1, \mathbf{h}_1 \vee \mathbf{h}_2)$, $P(+y \mid \mathbf{h}_2, \mathbf{h}_1 \vee \mathbf{h}_2)$, and $P(+y \mid \mathbf{h}_3, \mathbf{h}_2 \vee \mathbf{h}_3)$. From the first three parameters, $P(\mathbf{h}_i)$, $i = 1 \ldots 3$, can be recovered provided none of the prior probabilities is zero. The restriction that all prior probabilities are nonzero is quite reasonable. If the prior probability of some hypothesis were zero, there would be little reason to include that hypothesis in the model.

The other three parameters are equal to $P(+y \mid \mathbf{h}_1)$, $P(+y \mid \mathbf{h}_2)$, and $P(+y \mid \mathbf{h}_3)$, respectively, because $\mathbf{h}_i$ entails $\mathbf{h}_i \vee \mathbf{h}_j$. Thus $P(\mathbf{h}_i \mid +y)$ can be computed by Bayes rule:

$$P(\mathbf{h}_i \mid +y) = \frac{P(+y \mid \mathbf{h}_i) P(\mathbf{h}_i)}{\sum_{j=1}^{3} P(+y \mid \mathbf{h}_j) P(\mathbf{h}_j)}$$

In the case of type 1 similarity networks, this example indicates a general methodology for computing the posterior probability of each hypothesis. The general method is based on the proof of the following two theorems.

**Theorem 4 (restricted inferential completeness)** *Let $S$ be a type 1 similarity network of $P(h, u_1, \ldots, u_n)$ based on the connected cover $A_1, \ldots, A_k$ of the values of $h$. Let $\{v_1, \ldots, v_l\}$ be a subset of variables in $\{u_1, \ldots, u_n\}$ that satisfy relevant$(v_i, h)$. Then, the distribution $P(h, v_1, \ldots, v_l)$ can be computed from the parameters encoded in $S$ provided $P(\mathbf{h}_i) \neq 0$ for every value $\mathbf{h}_i$ of $h$.*

**Proof:** To show that the distribution $P(h, v_1, \ldots, v_l)$ can be computed from the parameters of $S$, we will show how to compute $P(h)$ and then we will show how to compute $P(v_1, \ldots, v_l \mid h)$. The product of these two probability distributions is equal to $P(h, v_1, \ldots, v_l)$.

For each hypothesis $\mathbf{h}_i$, let $\alpha_{ij}$ equal *INFER*($P(\mathbf{h}_i \mid [\![A_j]\!]), D_j$), where $A_j$ contains $\mathbf{h}_i$ and $D_j$ is the local network corresponding to $A_j$. The prior probability of each $\mathbf{h}_i$ is computed by solving the following set of linear equations:

$$P(\mathbf{h}_i) = \alpha_{ij} \cdot \sum_{\mathbf{h}_i \in A_j} P(\mathbf{h}_i), \quad \sum_{1}^{n} P(\mathbf{h}_i) = 1$$

In the previous section, we solved these equations and showed that the solution Equation 6 is unique provided $P(\mathbf{h}_i) \neq 0$ for all $\mathbf{h}_i$.

Due to the chaining rule, $P(v_1, \ldots, v_l \mid \mathbf{h}_i)$ can be factored as follows:

$$\begin{aligned} P(v_1, \ldots, v_l \mid \mathbf{h}_i) = \\ P(v_1 \mid \mathbf{h}_i) \cdot P(v_2 \mid v_1, \mathbf{h}_i) \ldots P(v_l \mid v_1, \ldots, v_{l-1}, \mathbf{h}_i) \end{aligned} \quad (7)$$

Thus, it suffices to show that for each variable $v_j$, $P(v_j \mid v_1, \ldots, v_{j-1}, \mathbf{h}_i)$ can be computed from the parameters encoded in $S$. Furthermore we can assume that the conditioning event is consistent, lest the entire product is zero, and the equality holds.

Let $D_i$ denote a local network in $S$, $A_i$ be the hypotheses associated with $D_i$, and $\mathbf{h}_i$ be a hypothesis in $A_i$. Each variable $v_j$ is depicted in some local network, because it satisfies *relevant*$(v_j, h)$. Let $A_i, A_{i+1}, \ldots, A_m$ be a path in the similarity hypergraph where $A_m$ is the only hyperedge on this path associated with a local network that depicts $v_j$ as a node. Such a path exists, because the similarity hypergraph is connected and $v_j$ is depicted in one of the local networks. If $v_j$ is depicted in $A_i$ then the entire path has one hyperedge: $A_i$.

Let $D_k$ be the local network associated with $A_k$ for $k = i+1 \ldots m$ and let $\mathbf{h}_{i+1}, \mathbf{h}_{i+2}, \ldots, \mathbf{h}_m$ be a sequence of hypotheses such that $\mathbf{h}_k \in A_{k-1} \cap A_k$. Due to the definition of type 1 similarity networks, since $v_j$ is not depicted in $D_k$ where $k < m$, $v_j$ is unrelated to $h$ given $[\![A_k]\!]$. Whenever $v_j$ is unrelated to $h$ given $[\![A_k]\!]$, $v_j$ is also mutually irrelevant to $h$ given $[\![A_k]\!]$. Thus,

$$\begin{aligned} P(v_j \mid v_1, \ldots, v_{j-1}, \mathbf{h}_{k-1}, [\![A_k]\!]) = \\ P(v_j \mid v_1, \ldots, v_{j-1}, \mathbf{h}_k, [\![A_k]\!]) \end{aligned}$$

Since $h$ implies $[\![A]\!]$ whenever $h \in A$ it follows that

$$P(v_j|v_1,\ldots,v_{j-1},\mathbf{h}_{k-1}) = P(v_j|v_1,\ldots,v_{j-1},\mathbf{h}_k)$$

This equation holds for every $k$ between $i+1$ and $m$, thus we obtain,

$$P(v_j|v_1,\ldots v_{j-1},\mathbf{h}_i) = P(v_j|v_1,\ldots v_{j-1},\mathbf{h}_m)$$

Furthermore,

$$P(v_j|v_1,\ldots v_{j-1},\mathbf{h}_m) = P(v_j|v'_1,\ldots v'_l,\mathbf{h}_m) \quad (8)$$

where $v'_1,\ldots,v'_l$ are the variables depicted in $D_m$ (a subset of $\{v_1,\ldots,v_{j-1}\}$).

If Equation (8) did not hold, then $related(v, v_j \mid [\![A_m]\!])$ would hold, where $v$ is some variable not appearing in $D_m$. Nonetheless, together with $related(v_j, h \mid [\![A_m]\!])$, which holds because $v_j$ is depicted in $D_m$, these two assertions imply by transitivity that $related(v, h \mid [\![A_m]\!])$ holds too, contradicting the fact that $v$ is assumed not to be included in $D_m$. Finally,

$$P(v_j|v'_1,\ldots v'_l,\mathbf{h}_m) = P(v_j|v'_1,\ldots v'_l,\mathbf{h}_m,[\![A_m]\!]), \quad (9)$$

because $\mathbf{h}_m$ logically implies the disjunction over all hypotheses in $A_m$.

The latter probability can be computed using *INFER* on the local network $D_m$. Thus, due to the equalities above, $P(v_j|v_1,\ldots,v_{j-1},\mathbf{h}_i)$ can be computed as needed. □

The above theorem shows that type 1 similarity networks are inferentially complete subject to the restriction that only features that help to discriminate between some hypotheses are included in the model and that all hypotheses which are included in the model have a probability greater than zero. Consequently, diagnostic completeness is guaranteed too.

**Theorem 5 (Diagnostic Completeness)** *Let $S$ be a type 1 similarity network of $P(h, u_1, \ldots, u_n)$. Then the conditional distribution $P(h \mid v_1, \ldots, v_m)$ can be computed from the parameters of $S$ for every subset $\{v_1, \ldots, v_m\}$ of $\{u_1, \ldots, u_n\}$ provided $P(\mathbf{h}_i) \neq 0$ for every value $\mathbf{h}_i$ of $h$.*

**Proof:** To compute $P(h \mid v_1,\ldots,v_m)$ observe that $P(h \mid v_1,\ldots,v_m) = P(h \mid v'_1,\ldots,v'_l)$ where $v'_1,\ldots,v'_l$ is the subset of variables in $v_1,\ldots,v_m$ that are relevant to $h$. Theorem 4 states that the joint distribution $P(h, v'_1, \ldots, v'_l)$ can be computed from the parameters of $S$. The conditional probability $P(h \mid v'_1,\ldots,v'_l)$ can be computed from this joint distribution. □

The above two theorems provide us with a naive computation of the posterior probability of each hypothesis. This computation does not take into account the fact that $P(h, v'_1, \ldots, v'_l)$ might be too large to be explicitly computed or stored as a table. Moreover, the computation suggested by these proofs ignore the crucial observation that, in practice, all local networks are often constructed according to a common order, say $h, v'_1, \ldots, v'_l$, which usually reflects cause-effect relations or time constrains.

When such a common ordering exists some computations become much easier. In particular, Equation 9 can be further developed. That is,

$$P(v_j|v'_1,\ldots v'_l,\mathbf{h}_m,[\![A_m]\!]) = P(v_j|\pi_{D_m}(v_j),[\![A_m]\!])$$
(10)

where $v'_1,\ldots,v'_l$ are the variables depicted in the local network $D_m$ and $\pi_{D_m}(v_j)$ are the parents of $v_j$ in $D_m$. Consequently, $P(v_j|v'_1,\ldots v'_l,\mathbf{h}_m,[\![A_m]\!])$ need not be computed using *INFER* as done in the proof. It is stored explicitly at node $v_j$ in the local network $D_m$.

Equation 10 defines a Bayesian network $M_i$ of $P(v_1,\ldots,v_m \mid \mathbf{h}_i)$, because, for each $v_j$, $\pi_{M_i}$ is set to be $v_j$ parents' set in $D_m$ excluding $h$, and the parameters associated with $v_j$ in $M_i$ are merely those associated with $v_j$ in $D_m$. The collection of these local networks, one network for each hypothesis $\mathbf{h}_i$, forms a structure that we have called *hypothesis-specific Bayesian multinet* of $P(h, v_1, \ldots, v_m)$ (Geiger and Heckerman, 1991).

**Definition** Let $P(h, u_1, \ldots, u_n)$ be a probability distribution and $\mathbf{h}_1, \ldots, \mathbf{h}_k$ be the values of $h$. A directed acyclic graph $D_i$ is called a *comprehensive local network of $P$ associated with $\mathbf{h}_i$* if $D_i$ is a Bayesian network of $P(u_1, \ldots, u_n \mid \mathbf{h}_i)$. The collection of $k$ comprehensive local networks is called an *hypothesis-specific Bayesian multinet* of $P$.

The algorithm below summarizes how to bulid this multinet. Afterwards, we shall show how to use it compute the posterior probability of each hypothesis. The first step below uses arc-reversal transformations in order to reorient all local networks according to a common construction order. This step is given for the purpose of completeness, namely, to enable the algorithm to process similarity networks that are not constructed according to a common construction order. In practice, however, this step usually is not needed, because similarity networks are constructed according to a common order of all relevant variables.

**Algorithm (Similarity network to Bayesian Multinet Conversion)**
**Input:** A type 1 similarity network $S$ of $P(h, u_1, \ldots, u_n)$ based on a connected cover $A_1, \ldots, A_k$ of the values of $h$.

**Output:** A hypothesis-specific Bayesian multinet of $P(h, v_1, \ldots, v_l)$ where each $v_i$ is depicted in some local network of $S$.

**Notation:**

- $M_i$ is the comprehensive local network associated with hypothesis $\mathbf{h_i}$
- $D_i$ is the local network associated with $A_i$
- $\pi_G(u)$ are the parents of $u$ in a graph $G$
- The conditional probability associated with node $u$ in $M_i$ is $P_{M_i}(u \mid \pi_{M_i}(u), \mathbf{h_i})$ and the condition-

al probability associated with node $u$ in $D_m$ is $P_{D_m}(u \mid \pi_{D_m}(u), \mathbf{h_m})$.

1. Reorient all local networks in $S$ according to a common construction order
2. For each $\mathbf{h_i}$ construct $M_i$ as follows
3.    For each $v_j$ taken in order $v_1, \ldots, v_l$
4.       Find a path $A_i, \ldots, A_m$ such that $\mathbf{h_i} \in A_i$ and $v_j$ is depicted only in $A_m$
5.       Set $\pi_{M_i}(v_j)$ to be $\pi_{D_m}(v_j) \setminus \{h\}$
6.       Set $P_{M_i}(v_j \mid \pi_{M_i}(v_j), \mathbf{h_i})$ to be $P_{D_m}(v_j \mid \pi_{D_m}(v_j), \mathbf{h_m})$

We now examine how the algorithm processes the similarity network $S$ in Figure 1. Because the node ordering $h, g, b, l$ is common to all local networks of $S$, the algorithm performs no arc reversals. Suppose the algorithm first builds the comprehensive local network $M_e$ for the hypothesis *executive*. Because $l$ appears in the local network for {*worker, executive*} with only $h$ as a parent, the algorithm makes $l$ a root note in $M_e$, and sets $P_{M_e}(l \mid executive)$ to be $P_{w \lor e}(l \mid executive)$, where $w \lor e$ denotes the local network for {*worker, executive*}. The local network for {*visitor, worker*} is the closest neighbor to the local network for {*worker, executive*} that depicts $g$ and $b$. Because the only parent of $g$ in the local network for {*visitor, worker*} is $h$, the algorithm makes $g$ a root node in $M_e$. Because $g$ and $h$ are the parents of $b$ in the local network for {*visitor, worker*}, the algorithm makes $g$ a parent of $b$ in $M_e$. The algorithm sets $P_{M_e}(g \mid executive)$ to be $P_{v \lor w}(g \mid worker)$ and $P_{M_e}(b \mid g, executive)$ to be $P_{v \lor w}(b \mid g, worker)$, where $v \lor w$ denotes the local network for {*visitor, worker*}. The algorithm constructs the comprehensive local networks for *worker*, *visitor* and *spy* similarly.

We can now use the inference algorithm stated below to compute the posterior probability of each hypothesis.

**Algorithm (Hypothesis-specific Bayesian-Multinet Inference)**

**Input:**

- A hypothesis-specific Bayesian multinet of $P(v_1, \ldots, v_m \mid h)$, where $M_i$ is the comprehensive local network associated with $\mathbf{h_i}$
- The apriori probability distribution $P(h)$
- Instances $\mathbf{v_1}, \ldots, \mathbf{v_l}$ for a set of variables $\{v_1, \ldots, v_l\} \subseteq \{v_1, \ldots, v_m\}$

**Output:** The posterior probability distribution $P(h \mid \mathbf{v_1}, \ldots, \mathbf{v_l})$

1. For each hypothesis $\mathbf{h_i}$

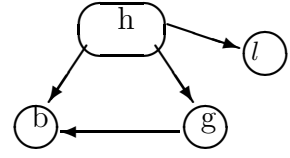

Figure 2: A Bayesian network representation of the secured-building story.

2.    $\beta_i = INFER(P(\mathbf{v_1}, \ldots, \mathbf{v_l} \mid \mathbf{h_i}), M_i)$
3. For each hypothesis $\mathbf{h}_i$
4.    Compute $P(\mathbf{h}_i \mid \mathbf{v_1}, \ldots, \mathbf{v_l}) = \frac{P(\mathbf{h}_i) \cdot \beta_i}{\sum_i P(\mathbf{h}_i) \cdot \beta_i}$

Line 2 is the normal computation performed by an inference algorithm for Bayesian networks. Line 4 encodes Bayes rule.

The advantage of computing $P(v_1, \ldots, v_l \mid \mathbf{h})$ via this algorithm versus using *INFER* on a Bayesian network of $P(v_1, \ldots, v_l, h)$ (see Figure 2) stems from hypothesis-specific independence assertions represented in some local networks, but not represented in Figure 2.

For example, suppose the guard of our secured-building problem sees a person wearing a badge (**b**) approach the building, but does not notice the person's gender or whether the person arrives in a limousine. Using the Bayesian network of Figure 2, *INFER* computes the posterior probability of each possible identification (*executive, worker, visitor, spy*) as follows:

$$P(h \mid \mathbf{b}) = k \cdot P(h) \cdot \sum_{\mathbf{g}} P(\mathbf{g} \mid h) \cdot P(\mathbf{b} \mid \mathbf{g}, h) \quad (11)$$

where $k$ is the normalizing constant that makes $P(h \mid \mathbf{b})$ sum to unity. Since the Bayesian network representing this problem does not encode any statement of conditional independence among $b$, $g$, and $h$, the above computation is done by any reasonable realization of *INFER*.

Alternatively, our inference algorithm computes the posterior probability of each hypothesis more efficiently, using the appropriate hypothesis-specific Bayesian multinet, as follows:

$$P(spy \mid \mathbf{b}) = k \cdot P(spy) \cdot P(\mathbf{b} \mid spy) \quad (12)$$

$$P(visitor \mid \mathbf{b}) = k \cdot P(visitor)$$
$$\cdot P(\mathbf{b} \mid visitor) \quad (13)$$

$$P(worker \mid \mathbf{b}) = k \cdot P(worker)$$
$$\cdot \sum_{\mathbf{g}} P(\mathbf{g} \mid [\![worker, executive]\!]) \quad (14)$$
$$\cdot P(\mathbf{b} \mid \mathbf{g}, [\![worker, executive]\!])$$

$$P(executive \mid \mathbf{b}) = k \cdot P(executive)$$
$$\cdot \sum_{\mathbf{g}} P(\mathbf{g} \mid [\![worker, executive]\!]) \quad (15)$$
$$\cdot P(\mathbf{b} \mid \mathbf{g}, [\![worker, executive]\!])$$

Equations 12 and 13 take advantage of hypothesis-specific independence. In particular, the two equations incorporate the fact that $g$ and $b$ are conditionally independent given $h = spy$ and $h = visitor$, respectively. Thus, we do not have to sum over the variable gender as we do when using the Bayesian network of Figure 2 (Equation 11). These savings are achieved by our inference algorithm for Bayesian multinets, because the algorithm applies *INFER* on a local network that encodes this independence information.

Note that the sums in Equations 14 and 15 are equal due to subset independence. Our inference algorithm, however, does not use this fact or any other assertion of subset independence. The inference algorithm for strictly positive distributions, on the other hand, computes this sum only once.

## 6 OPEN PROBLEMS

We conjecture that every type 2 similarity network also satisfies Theorems 4 and 5, and that the above algorithm is applicable to these networks as well. We further believe that there exists an inference algorithm that uses both subset independence as well as hypothesis-specific independence, even when the distributions are not strictly positive.

### Acknowledgments

The second author thanks Mark Peot for suggesting an early version of the inference algorithm for strictly positive distributions. This research was supported in part by the fund for the promotion of research in the Technion.